*Genome analysis*

# Subtype-Former: a deep learning approach for cancer subtype discovery with multi-omics data

Hai Yang[1], Yuhang Sheng[1], Yi Jiang[2], Xiaoyang Fang[3], Dongdong Li[1], Jing Zhang[1], Zhe Wang[1*]

[1]Department of Computer Science and Engineering, East China University of Science and Technology, Shanghai 200237, China,

[2]Department of Epidemiology and Biostatistics, Tongji Medical College, Huazhong University of Science and Technology, Hubei, 430030, PR China,

[3]Jacobs Technion-Cornell Institute, Cornell Tech, Cornell University, New York, 10044, USA

**Abstract**
**Motivation:** Cancer is heterogeneous, affecting the precise approach to personalized treatment. Accurate subtyping can lead to better survival rates for cancer patients. High-throughput technologies provide multiple omics data for cancer subtyping. However, precise cancer subtyping remains challenging due to the large amount and high dimensionality of omics data.
**Results:** This study proposed Subtype-Former, a deep learning method based on MLP and Transformer Block, to extract the low-dimensional representation of the multi-omics data. K-means and Consensus Clustering are also used to achieve accurate subtyping results. We compared Subtype- Former with the other state-of-the-art subtyping methods across the TCGA 10 cancer types. We found that Subtype-Former can perform better on the benchmark datasets of more than 5000 tumors based on the survival analysis. In addition, Subtype-Former also achieved outstanding results in pan-cancer subtyping, which can help analyze the commonalities and differences across various cancer types at the molecular level. Finally, we applied Subtype-Former to the TCGA 10 types of cancers. We identified 50 essential biomarkers, which can be used to study targeted cancer drugs and promote the development of cancer treatments in the era of precision medicine.
**Availability:** All the data is available in The Cancer Genome Atlas (https://www.cancer.gov/tcga). The source codes of Subtype-Former are available at https://github.com/haiyangLab/Subtype-Former.
**Contact:** wangzhe@ecust.edu.cn
**Supplementary information:** Supplementary data are available at *Bioinformatics* online.

## 1 Introduction

Cancer is a group of severe genomic diseases that threaten human life and health (Ushijima *et al.*, 2021). Modern precision medicine requires targeted genomic therapies in cancer treatment for different patients (Pleasance *et al.*, 2020). However, the challenge is the significant heterogeneity between patients, tumor types, tumor cells, and clones (Liu *et al.*, 2021). By using genomic data from samples, molecular subtyping allows for a more detailed classification of tumors at the molecular level, leading to subtypes with similar biological characteristics or survival times and treatment effects in corresponding cancer. The findings help to explain the mechanisms of cancer pathogenesis and thus develop specific treatment regimens to improve patient survival, thereby enhancing the overall understanding of tumorigenesis in different tissues and advancing the development of cancer genomics. However, this task remains a significant challenge in cancer research due to the diversity, complexity, and specificity of cancer genomic data (Vitale *et al.*, 2021).

Molecular subtyping that focuses on individual omics data may ignore commonalities between the multi-omics data of cancer patients, thereby limiting the accuracy of the results (Ramazzotti *et al.*, 2018). Cancer subtyping based on multi-omics data integration can use complementary information to describe cancer patients. With the development of high-throughput sequencing technology, an increasing number of international organizations such as The Cancer Genome Atlas (TCGA) (The Cancer Genome Atlas Research Network *et al.*, 2013), International Cancer Genome Consortium (ICGC) (The International Cancer Genome Consortium, 2010), Pan-Cancer Analysis of Whole Genomes (PCAWG) (Pleasance *et al.*, 2020) have collected multiple types of cancer omics data (such as genomics, epigenomics, transcriptomics, proteomics) (Lu *et al.*, 2021). Benefit from these projects, efficient multi-omics subtyping methods can be developed to understand the occurrence and development of cancer and propose effective treatments. However, this task is not straightforward, as the increase in the number of omics makes the data too dimensional and increases confusion across platforms, making integrating multi-omics difficult (Chauvel *et al.*, 2020, 202).

Several approaches have been proposed to achieve accurately molecular tumor subtyping (Li *et al.*, 2016; Zhao *et al.*, 2019). Depending on the model's input, we classify all methods into single-input (SI) and multi-input (MI) methods. The SI methods combine multi-omics data into a large input matrix and perform a straightforward clustering approach. For instance, MCCA (Witten and Tibshirani, 2009) transfers the multi-omics input into the lower dimensional space to maximize the correlation between features and then perform the clustering. moCluster (Meng *et al.*, 2019, 2016) transforms multi-omics learning features to the same scale through a low-dimensional representation of those features and computes the clustering results. LRACluster (Wu *et al.*, 2015) ties together multiple heterogeneous omics data for each sample by modeling the distribution of numerical, count, and discrete features. Subtype-WESLR (Song *et al.*, 2022) projects each sample feature profile into a corresponding common latent subspace that maintains the local structure and is consistent with the integrated clustering to identify cancer subtypes.

The MI methods take multiple data sources directly as input and complete the subtyping task. Among them, iClusterBayes (Mo *et al.*, 2018) uses the Bayesian latent variable regression model to integrate multi-omics data by projecting them into a common low-dimensional integration space. SNF (Wang *et al.*, 2014) constructs a similarity network between samples, using a message passing process to update the similarity weights of multiple similarity networks iteratively. NEMO (Rappoport and Shamir, 2019) proposes a neighborhood-based multi-group clustering algorithm based on similarity networks and copes well with missing omics data. Subtype-GAN (Yang *et al.*, 2021) based on deep neural networks. Adversarial learning is used to reduce the dimensionality of each omics data individually, and then the low-dimensional data is stitched together for clustering. PINS (Nguyen *et al.*, 2017, 2019) clusters each omics dataset using the perturbation clustering method and constructs a connectivity matrix, which is then combined and assembled into a similarity matrix. The addition of noise makes the clustering results more robust.

Molecular subtyping focusing only on a single cancer type has the potential to overlook the commonalities in the omics data of patients with different cancer types, resulting in molecular mechanisms that are difficult to explain in some cancers and limiting the development of new therapeutic modalities. As different types of tumors share commonalities at the molecular level, such as genome and transcriptome, pan-cancer subtyping studies can analyze the commonalities and differences between different cancer types at the molecular level, thereby understanding the links and differences between subtypes across cancer types and extending established treatments to rare cancer types with similar omics data. With the rapid accumulation of omics data on different cancer types, studies have been conducted on pan-cancer molecular subtyping across tissue samples for more accurate cancer diagnosis and treatment. However, due to the difficulties of large sample size and high dimensionality of multi-omics data faced by pan-cancer subtyping, relevant bioinformatic methods are still scarce. It has been shown that deep learning methods have significant advantages over traditional pan-cancer subtyping methods (Duan *et al.*, 2021).

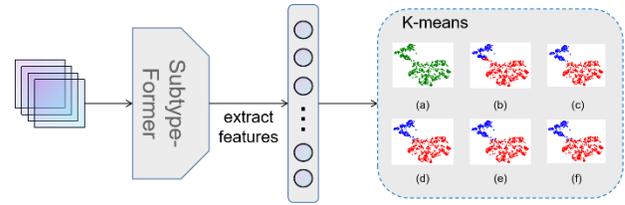

**Fig. 1** Flow chart of Subtype-Former. It contains two parts: extracting low-dimensional hidden features from multi-omics cancer data using the MLP-Transformer network and K-means clustering for cancer subtyping.

Accurate modeling and subtyping of complex multi-omics data from individual cancers remain challenging. Moreover, on the task of pan-cancer subtyping with colossal sample size, the excessive complexity of the statistical model may cause the subtyping task being unable to be completed in an acceptable time. Since Deep learning has great potential to accurately model complex omics data, in this paper, we proposed Subtype-Former, a novel unsupervised learning model for cancer subtyping. Subtype-Former innovatively integrated multiple layers of MLPs and transformer blocks to form an unsupervised network, which improves the ability to represent and reconstruct complex omics data in hidden space. We analyzed the ten cancer types with the largest sample sizes in the TCGA portal. The comparison results show that Subtype-Former is faster and has superior performance in cancer subtyping tasks compared to the current state-of-the-art subtyping approaches. We also conducted the pan-cancer subtyping analysis across TCGA 32 cancer types (Hoadley *et al.*, 2018). Subtype-Former completed the analysis in 302 seconds and yielded 27 subtypes with significant survival differences (p-value < 1e-6). Finally, we explored the subtypes on the KIRC dataset in-depth and found three critical biomarkers in KIRC with the Random Forest algorithm. Meanwhile, we used the t-SNE algorithm to reduce high-dimensional features to 2 dimensions to visualize their data characteristics. We found that the subtypes obtained by Subtype-Former had apparent differences in the t-SNE plots. Moreover, the results showed significant survival differences between the subtypes of KIRC. Overall, our proposed deep learning-based Subtype-Former can perform the task of subtyping on each cancer type for multi-omics data integration, and the task of pan-cancer subtyping and the results can contribute to our understanding of the heterogeneity within cancers at the molecular level.

## 2 Methods

We used the multi-omics data across TCGA ten cancer types as input to the Subtype-Former. The output of the method was the clustered labels for each sample. When the numbers of subtypes are specified, the hidden features were clustered using K-means to obtain the corresponding labels. The cluster number and the subtype labels were automatically obtained with the consensus clustering, while the subtype number was not specified. The working flow of Subtype-Former is shown in Figure 1.

### 2.1 Data pre-processing and feature selection

This study focuses on ten cancer datasets from the TCGA database with the largest sample size (including BRCA, UCEC, HNSC, THCA, LUAD, KIRC, PRAD, LUSC SKCM, STAD). More than 50% of all TCGA cancer samples were included. The sufficient data allowed both Subtype-Former and its comparative methods to obtain stable subtyping results. According to the previous study, we collected four omics corresponding to all samples: copy number, mRNA, miRNA, and DNA methylation to extract features (Yang *et al.*, 2021).

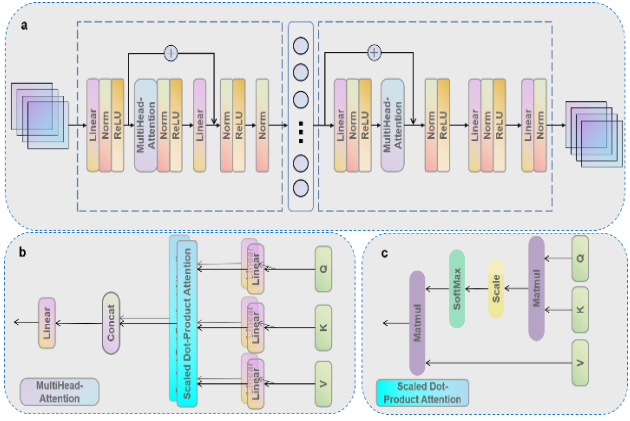

**Fig. 2 Subtype-Former network structure.** (A) The network architecture of Subtype-Former, including Linear, Norm, ReLU, Residual, and MultiHead-Attention modules. The encoder and decoder are almost symmetric. (B) Detailed architecture of the MultiHead-Attention module in Subtype-Former. (C) The specific structure of the Scaled Dot-product Attention module used in the MultiHead-Attention module.

First, we pre-processed the raw data for the four omics. We extracted data for each cancer patient sample for its corresponding four omics and used the experience from previous studies for feature selection. Specifically, the model input to the Subtype-Former included 3105 copy number values, 3217 DNA methylation values, 383 miRNA values, and 3139 RNA values. Previous studies have shown that the weight of the contribution of each omics data to the subtype results tends to be contrasted across different cancer types. The Subtype-Former network is a powerful non-linear representation with attention mechanism modules that can automatically focus on critical features in multi-omics data to characterize complex data and automatically obtain reasonable subtype results.

### 2.2 The neural network of the Subtype-Former approach

Unlike the traditional Transformer's input in natural language processing and computer vision, the input data of Subtype-Former is high-dimensional features extracted from the multi-omics data. Hence, we innovatively integrated the multilayer perceptron (MLP) with Transformer modules to form an unsupervised network that enhances the ability of the hidden layer to represent and reconstruct complex omics data via the Transformer's attention mechanism. The encoder part of Subtype-Former uses MLP to reduce complex high-dimensional data into lower dimensions. In contrast, the Subtype-Former's decoder layer network decodes low-dimensional data into high-dimensional features with the same dimensionality as the input data. Subtype-Former uses Transformer blocks in both encoder and decoder to enhance the model's power.

Specifically, for the input sample xi, the Subtype-Former's encoder compresses it into the intermediate layer of the network, while the decoder extends the low-dimensional representation mi to the same size as the input sample $x_i$:

$$m_i = f_{encoder}(x_i), \ x_i = f_{decoder}(m_i) \quad (1)$$

The network contains an input layer, an output layer, and several hidden layers. Between each layer, we use ReLU as the activation function of the network to avoid the problem of gradient disappearance in the network:

$$\operatorname{Re}LU(x) = \max(0, x) \quad (2)$$

Also, between each layer of the network, we use Layer-Normalization to prevent overfitting problems:

$$\mu_L = \frac{1}{d}\sum_{i=1}^{d} x_i$$
$$\sigma_L^2 = \frac{1}{d}\sum_{i=1}^{m_k d}(x_i - \mu_L)^2 \quad (3)$$
$$LN(x) = \gamma \frac{x - \mu_L}{\sqrt{\sigma_L^2 + \varepsilon}} + \beta$$

where $\mu_L$ and $\sigma_L$ are the mean and variance of all features of a sample in a layer of the network, respectively, and d is the dimensionality of the sample in that layer. The input x is adjusted by $\gamma$ and $\beta$ to conform to a normal distribution. The parameter $\varepsilon$ is introduced to prevent variance from going to zero. In addition, a multi-head attention mechanism is used in both the encoder and decoder to improve the network's ability to extract hidden features from the data. The expression of the self-attention function is as follows:

$$Attention(Q, K, V) = soft\max(\frac{QK^T}{\sqrt{d_k}})V \quad (4)$$

where $Q, K, V$ are projections of the input $x$, and they are all matrices. $Q, K, V$ correspond to query, key, and value, respectively, and their dimensions are $d_k, d_k, d_v$. The Key and value represent a set of key-value pairs, and the query is the ordinal number corresponding to the key-value pair. We compute the dot product of query and key and divide it by $\sqrt{d_k}$ and obtain the weight on value by using the Softmax function. In multi-head attention, we compute the attention output of the $h$ headers in parallel using the different projection parameters learned. Parallel computation of multiple heads can significantly reduce the computation time of the multi-head attention layer and improve computational efficiency. The Multi-head attention allows the model to focus on the expressed information in different heads simultaneously:

$$MultiHead(Q, K, V) = Concat(head_1, ...head_h)W^O$$
$$head_i = Attention(QW_i^Q, KW_i^K, VW_i^V) \quad (5)$$

where $W_i^Q \in \mathbb{R}^{d*d_k}$, $W_i^K \in \mathbb{R}^{d*d_k}$, $W_i^V \in \mathbb{R}^{d*d_v}$, $W^O \in \mathbb{R}^{h*d*d_v}$. In this work, $h = 2, d_k = d_v = 64$. The specific network structure is shown in Figure 2.

Finally, the network reconstructed the features with the decoder, and we use Mean-Square Error (MSE) as the loss function of the Subtype-Former network:

$$\ell_{MSE} = \frac{1}{N}\sum_{i=1}^{N}(x_i - x_i)^2 \quad (6)$$

where N represents the dimensionality of the input data, $x_i$ is the original sample, and $x_i$ is the data transformed by the Subtype-Former to have the same dimensionality as the original sample.

The description of the specific parameters in the network can be found in Supplementary Material 1. The parameters of the Subtype-Former are updated until all epochs are completed. After all training, the resulting hidden layer features were input into the clustering module. We used the K-means method as the clustering method to aggregate the middle layer vectors into the specified K classes.

### 2.3 Datasets and comparison algorithms

We used the TCGA dataset as input to the deep model, which is available at The Cancer Genome Atlas (https://www.cancer.gov/tcga). We presented the case of data in Supplementary Table S2. In addition, we used clinical data of these samples as the key to assessing the subtyping performance. We designed two test scenarios to evaluate Subtype-Former on the benchmark sets comprehensively.

Firstly, we evaluated the performance of the Subtype-Former on individual cancer types. We selected 5316 samples from each of the ten cancer types with the largest sample sizes of all TCGA cancer types to construct the test dataset (including 1031 BRCA tumors, 510 UCEC tumors, 506 HNSC tumors, 494 THCA tumors, 490 LUAD tumors, 488 KIRC tumors, 484 PRAD tumors, 460 LUSC tumors, 446 SKCM tumors, 407 STAD tumors). We used its four omics data for each sample, including copy number, methylation, miRNA, and mRNA. We selected nine of the latest cancer subtyping methods as comparison algorithms, including LRACluster, MCCA, moCluster, Subtype-GAN, SNF, NEMO, CIMLR, iClusterPlus, and PINS. These methods include both traditional and deep learning algorithms. To ensure a fair comparison, the parameter settings of all procedures followed the developers' recommendations (see Supplementary Material 2 for details).

We selected 7862 samples on 32 TCGA cancers for the pan-cancer subtyping task. Four types of omics data, copy number, methylation, miRNA, mRNA, and clinical data, were available to form the pan-cancer test dataset. The omics features of these samples were extracted in the same way in the previous test scenario. The following comparison algorithms were chosen: LRACluster, MCCA, moCluster, Subtype-GAN, SNF, NEMO, CIMLR, iClusterPlus, and PINS.

Follow the precedent (Rappoport and Shamir, 2018), we used the P-value of the significant difference in survival analysis in both test scenarios to evaluate all algorithms. To avoid the log-rank P-value being insufficiently accurate due to too substantial differences in sample size across subtypes, the P-value was calculated using empirical estimates based on the permutation test. In the single cancer test scenario, we also select the same set of clinical parameters for all cancers: gender, tumor progression (pathology T), lymph node cancer (pathology N), metastasis (pathology M), total progression (pathology stage) and age at initial diagnosis.

## 3 Results

### 3.1 The performance analysis of the subtyping methods across 10 TCGA cancer types

We compared the Subtype-Former with the nine most recent cancer subtyping methods. Based on the survival analysis, we calculated the empirical P-values of the log-rank test for all the methods (Fig. 3A, Fig. 3C). Also, based on the enrichment analysis (chi-square test and Kruskal-Wallis test), we reported the number of significant clinical parameters across each cancer type (Fig. 3B, Fig. 3D). To avoid the influence of clusters k on the results, we set a reasonable number of clusters k on each dataset based on previous studies (Akbani *et al.*, 2015; Berger *et al.*, 2018; The Cancer Genome Atlas Research Network, 2014, 2013; The Cancer Genome Atlas Research Network and Levine, 2013; Chung *et al.*, 2004; The Cancer Genome Atlas Research Network, 2012; Abeshouse *et al.*, 2015;

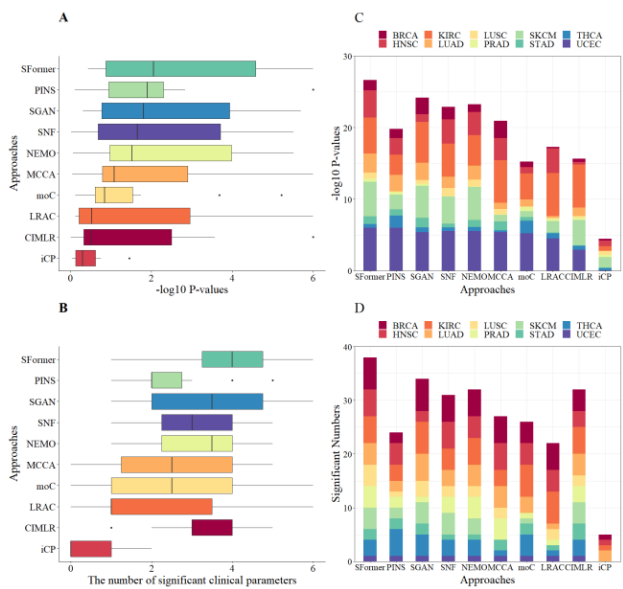

**Fig. 3 Performance of Subtype-Former compared with other methods (including PINS, Subtype-GAN, SNF, NEMO, MCCA, moCluster, LRACluster, CIMLR, iCluster-Plus).** (A) -logP-value of ten methods. (B) The number of significant parameters (including sex, tumor progression (pathological T), lymph node status (pathological M), total progression (pathological stage), and age). (C) -logP-value for stacking of results on ten cancers. (D) The number of significant parameters stacked across ten cancers.

Agrawal *et al.*, 2014; Liu *et al.*, 2018) (see Table S3 in the Supplementary Material for choosing the number of clusters k). We found that Subtype-Former achieved the best results relative to all comparison methods (-log10 P-value mean of 2.66 and median of 2.06; the number of significant clinical parameters mean of 3.8 and median of 4). Specifically, Subtype-Former achieved the most significant survival differences across the UCEC, HNSC, LUAD, and SKCM (-log10 P-value of 6 on UCEC, 3.78 on HNSC, 2.67 on LUAD, and 4.85 on SKCM) and achieved significant results on 6 of 10 cancers (empirical log-rank test P-value < 0.05). Also, for the indicator of clinical parameters, Subtype-Former achieved the best performance on BRCA, UCEC, HNSC, PRAD, LUSC, and SKCM (6,1,5,4,4,4 respectively). Overall, Subtype-Former was either superior to or competitive with other methods in survival analysis versus clinically significant parameter analysis (see Supplementary Material S3 for specific results). We also add two metrics, NMI and ARI, to evaluate the clustering results and examine the performance of Subtype-Former (see Supplementary Note 7 for details). Due to the peculiarity of deep learning and the absence of huge matrix calculation, Subtype-Former also performs well in computational efficiency. We also record the running times of all methods (Supplementary Table S4).

To further analyze the differences between the results of the cancer subtyping methods, we performed a Friedman analysis of all the methods (Fig. 4). We observed that Subtype-Former significantly outperforms iCluster-Plus (P-value < 0.05) on 10 datasets, but not always better than all other methods. In addition, we performed hierarchical clustering of the Subtype-Former with other methods (see Supplementary Note 4 for details). We found that hierarchical clustering results were stable, and the better-performing methods were always clustered into similar categories. Subtype-Former, NEMO, and SNF were always grouped into the same category, which suggested that our approach achieved stable and excellent performances across the ten tumor types.

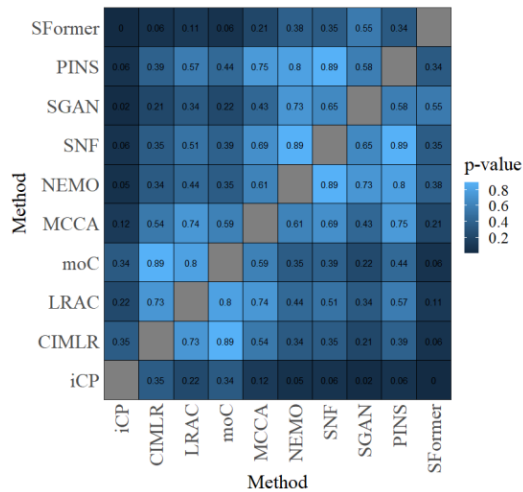

**Fig. 4 Friedman test to show the differences among all methods. Subtype-Former is significantly better than iClusterPlus on ten datasets (p-value < 0.5).**

### 3.2 Automatically confirm the number of subtypes with Subtype-Former

In the first section of this chapter, we used the fixed number of clusters for Subtype-Former. However, the number of subtypes of some tumor types cannot yet be entirely determined, especially for some rare tumors. Fortunately, Subtype-Former also supports automatic selection of the appropriate number of cancer subtypes to suit more application scenarios. We used the consensus clustering method on the hidden layer of the unsupervised MLP-transformers network to obtain the number of clusters and perform subtyping simultaneously. Next, by consensus clustering, Subtype-Former obtained the number of clusters automatically selected for the ten cancers (see Supplementary Material S5 for details).

By calculating the -log10 P-values and the significant clinical parameters, we found that the performance of Subtype-Former with Consensus Cluster (Subtype-Former-CC) was competitive compared with the previous benchmark results in terms of -log10 P-value (the mean of 2.48, the median of 2.00) and the number of significant clinical parameters (the mean of 3.3, the median of 3.5). Hence, we consider that Subtype-Former-CC also obtains reliable results across the ten types of cancer.

### 3.3 Performance analysis of Subtype-Former on the Pan-Cancer profiles

In the study of cancer genomics, Pan-Cancer subtyping is also essential. The TCGA launched the Pan-Cancer Analysis Project (PCAP) in 2012 to find similar cancer subtypes in different tumors at the molecular level and extend effective treatments to other cancer subtypes with similar genomic profiles. Subsequently, Hoadley et al. (Hoadley *et al.*, 2018) revealed that 33 types of cancer could be grouped into 28 categories in a pan-cancer study of more than 10,000 samples in TCGA. Based on this study, we selected 7862 samples with all four omics data (32 cancers in total, excluding LAML) to demonstrate that the Subtype-Former is equally valid for this task. The feature values also include the CNV, DNA methylation data, miRNA data, and RNA data.

Benefit from the advantage that deep learning can accurately model a large number of samples, we successfully separated the pan-cancer tumors into 27 categories with Subtype-Former. We labeled these 27 clusters as C1~C27 and used heat maps on the hidden factors generated by Subtype-

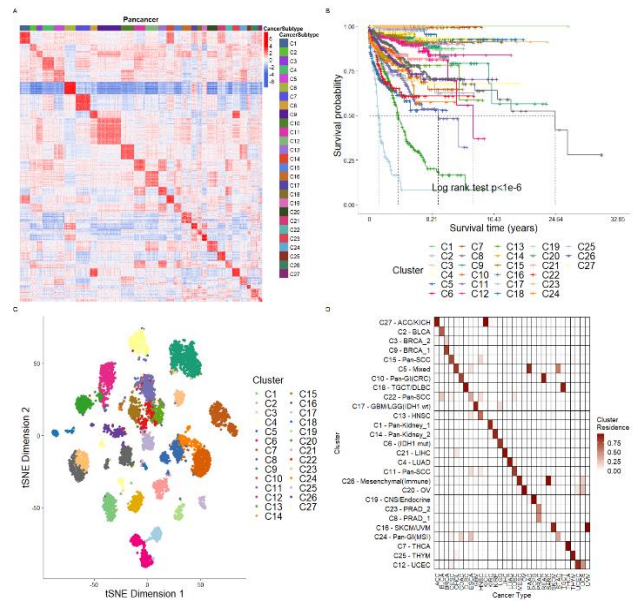

**Fig. 5 Analysis of Pan-Cancer subtyping results of Subtype-Former:** (A) Heatmap of the correlations of the hidden factors (64 dimensions) between all samples. The x-axis and the y-axis are both pan-cancer subtypes. Red mains high correlation and blue is a low correlation. (B) Kaplan-Meier survival analysis plot of Pan-Cancer with significant survival differences between clusters. (C) Results of dimensionality reduction using t-SNE of Subtype-Former. (D) Correspondence between Pan-Cancer clusters and specific cancers.

Former to visualize the similarities of subtypes (Fig. 5A). We observed that each subtype shows a clear boundary. Simultaneously, some clusters have a clear correlation (e.g., C1 and C14, C8 and C23), suggesting that some clusters are significantly correlated while others are different. To demonstrate that C1~C27 have significant survival differences, we performed the Kaplan-Meier survival analysis on 7862 samples (Fig. 5B). The 27 clusters exhibited significant survival differences (empirical P-value < 1e-6). Across all clusters, the worst prognosis for survival was C17 (mainly GMM/LGG patients), while the best prognosis was C19 (mainly PCPG patients), consistent with our knowledge of these cancers (Adamson *et al.*, 2009; Thosani *et al.*, 2013). To get a more intuitive illustration of the differences between the different clusters, we used t-SNE to downscale the middle layer of the Subtype-Former network to two dimensions to observe the aggregation of the data across the subtypes. We used the labels obtained from Subtype-Former for t-SNE visualization (Fig. 5C) and observed that samples from the same cancer type were always classified into one cluster, while samples from different cancer types were usually significantly separated. Also, the distance between samples from related types of cancer is generally close (e.g., C8 and C23), which remains consistent with previous discoveries.

We found that among all the 27 clusters identified by Subtype-Former, 15 were dominated by one cancer (C1: KIRC, C2: BLCA, C3: BRCA, C4: LUAD, C6: LGG, C7: THCA, C8: PRAD, C9: BRCA, C12: UCEC, C13: HNSC, C14: KIRP, C19: PCPG, C21: LIHC, C23: PRAD, C25: THYM), the remaining clusters are composed of several cancers (C11/C15/C22: mainly consisting of squamous cell carcinomas including BLCA, CESC, ESCA, HNSC, LUSC; C5/C10/C24: mainly composed of cancers of gastrointestinal cancer including CHOL, COAD, ESCA, READ, STAD, PAAD; C16 includes melanomas such as SKCM and UVM; C17 includes brain cancers like GBM and LGG; C18 includes DLBC and TGCT; C20 includes female cancers such as OV, UCEC, UCS; C26 includes melanomas such as SARC, MESO, SKCM; C27 includes suprarenal cell carcinomas such as ACC, KICH. We named the clusters to clarify the relationship between cancers and clusters (Fig. 5D).

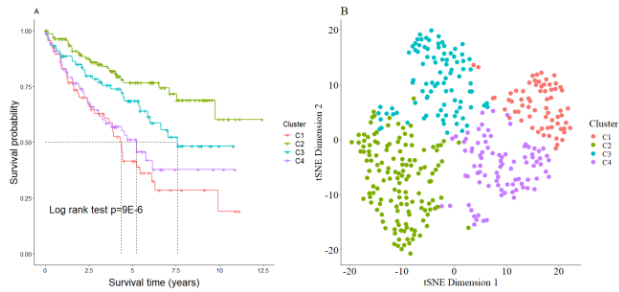

**Fig. 6 Subtype-Former applied to TCGA KIRC tumors.** (A) The Kaplan-Meier survival analysis plot on the KIRC profiles. (B) The Subtype-Former hidden factors (64 dimensions) were visualized using t-SNE.

We also performed the pan-cancer subtyping task with the other methods which support Pan-Cancer dataset. All of them completed the Pan-Cancer subtyping task with significant results (empirical P-value < 1e-6). However, there were substantial differences in the running time of individual approaches. Due to the large sample size of the Pan-Cancer benchmark set, most established methods cannot give reliable clustering results in a short time. The deep learning-based Subtype-Former can obtain a reasonable subtyping scheme fast and accurately. By comparing the running time of the different methods across the 7862 samples, we found that Subtype-Former was second only to Subtype-GAN for the pan-cancer task and significantly faster than the other non-deep learning methods (see Table S4 in Supplementary Material), which is consistent with the conclusion from the previous studies. In terms of operational efficiency, deep learning methods such as Subtype-Former are more suitable for increasing volumes of data while also delivering consistent and reliable results.

### 3.4 Case study on the KIRC

Finally, we applied Subtype-Former to the 387 TCGA KIRC patients for a specific analysis of the subtyping results. Subtype-Former obtained four subtypes of KIRC (denoted as C1-C4) through the consensus cluster. With the Kaplan-Meier survival analysis (Fig. 6A), we found significant survival differences between the four subtypes of KIRC (empirical P-Value = 9e-6), as well as a sufficient number of clinical indicators enriched for the different subtypes, indicating that Subtype-Former can distinguish the subtypes of KIRC tumors well without any prior knowledge (such as the number of subtypes and the age of patient). Next, to explore the difference between subtypes of KIRC patients, we used the t-SNE plot to downscale the hidden factors of Subtype-Former to 2 dimensions for visualization (Fig. 6B). We found that patients from the same subtype are always grouped. The difference between the low-dimensional features of patients from different subtypes is evident. The four subtypes of KIRC patients have clear boundaries in the t-SNE figure.

Also, to enhance the interpretation and understanding of the Subtype-Former subtyping results. We found the three most critical biomarkers of the subtypes of KIRC: CDKN2B-AS1, GIPC2, and HAO2. The literature study found that CDKN2B-AS1 (Dasgupta *et al.*, 2020) is an essential oncogenic lncRNA that plays a vital role in renal carcinogenesis. Overexpression of CDKN2B-AS1 was positively correlated with poorer overall survival in RCC patients. GIPC2 (Liu *et al.*, 2020) was significantly associated with prognosis in RCC patients. Low expression of GIPC2 tended to imply lower survival in RCC patients, with hypermethylation and lower expression of GIPC2 leading to low survival rates. In addition, HAO2 (Xiao *et al.*, 2019, 2) inhibits malignant KIRC by promoting lipolytic metabolic processes and may be an effective molecular marker and treatment for RCC, with low HAO2 associated with shorter overall survival (OS) and shorter disease-free survival (DFS). These literature studies suggested that the Subtype-Former clustering results on the KIRC dataset were reliable. Due to the complexity of deep learning, people usually obtain the interpretation of methods through other kinds of algorithms (such as supervised learning). We use a tree-based approach, Random Forest, to explain the relationship between omics data and subtyping results of the model. The input of it corresponds to the input of Subtype-Former. Output is the most important biomarker of cancer. Similarly, we performed the Random Forest on nine other TCGA tumor datasets and identified corresponding biomarkers of the subtypes (Supplementary Material S6).

## 4 Discussion

With the rapid development of high-throughput technologies, multi-omics integration methods have emerged in cancer subtyping. This study presents Subtype-Former, a deep learning method based on the unsupervised MLP-Transformer-network for cancer subtyping. We compared Subtype-Former's performance with nine state-of-the-art methods on the ten single-cancer datasets and pan-cancer datasets of TCGA. We found that Subtype-Former can achieve superior or comparable performance on the benchmark dataset (-log10 P-value mean of 2.66 and median of 2.06; the number of significant clinical parameters mean of 3.8 and median of 4). We also compared the performance of 10 methods on a pan-cancer dataset. Subtype-Former obtained substantial and consistent results with most other methods. Finally, we applied Subtype-Former to 387 samples from KIRC for subtyping and used the Random Forest algorithm to find three biomarkers (CDKN2B -AS1, GIPC2, HAO2). The literature study identified these three genes to be significantly associated with survival in KIRC patients. We similarly identified several biomarkers for the remaining nine cancers, and previous studies could support these 50 biomarkers. If the number of clusters k is not specified, Subtype-Former also supports the automatic confirmation of the number of clusters by consensus clustering. The subtyping results for the automatic selection of clusters do not differ much from the results for the specified number of clusters. Subtype-Former's consensus clustering gives more reliable and stable results than the previous deep learning method Subtype-GAN.

To illustrate the differences more visually between subtyping results, we used t-SNE to downscale the middle layer of the Subtype-Former network to two dimensions to observe the data aggregation across the subtypes. T-SNE dimension reduction was performed for both KIRC and pan-cancer subtypes. Comparing individual cancers with pan-cancers, we found that the different subtypes of individual cancers are closer together after dimensionality reduction, and it is challenging to find clear boundaries. However, after dimensionality reduction of the pan-cancer subtypes, the distance between the subtypes occurring in different tissues is far apart. In contrast, the space in the same tissue is relatively close, consistent with the conclusions of previous studies.

Transformer network has achieved significant improvements in several studies in biomedicine, demonstrating the power of deep learning. Recently, DeepMind presented the AlphaFold2 framework (Tunyasuvunakool *et al.*, 2021) which improves the accuracy of protein structure prediction to 98.5% with an improved Transformer network. They also came up with a generalist agent named Gato, which is working as a multi-modal, multi-task, multi-embodiment policy with Transformer blocks (Reed *et al.*,

*Subtype-Former*

2022). In these studies, the Transformer works in a supervised mode. Subtype-Former was developed as an unsupervised Transformer network, which combined with the non-linear representation capabilities of MLP and a multi-head attention mechanism to achieve an accurate low-dimensional representation of complex multi-omics data. As far as we know, Subtype-Former is the first method to apply the Transformer to the field of cancer subtyping and has achieved excellent performance.

Subtype-Former and Subtype-GAN are both unsupervised deep learning methods. The main difference between the two methods is whether the model is an SI model or an MI model. At the same time, the performance of MCCA and PINS algorithms also works well with the SI strategy and the MI strategy. The SI strategy of Subtype-Former simplifies the structure of the model (from a multiple-input, multiple-output network to a single-input, single-output network), thus making it easier to incorporate the latest network structures from the deep learning community and improve the model's power. Subtype-GAN uses an MI strategy, but the contribution analysis for different cancers shows that different omics data's contribution to the results varies in cancer types (also confirmed by CIMLR). In contrast, the multiple-input model considers different omics data to have fixed weights, which may be inappropriate. We consider that both single-input and multiple-input methods are suitable for molecular subtyping for multi-omics integration. The performance of subtyping algorithms can be improved when the methods themselves can assign higher weights to essential features. The multi-head attention mechanism in Subtype-Former is inspired by human attention, selecting the most critical features. Compared to deep adversarial networks, it can more accurately find the essential biometric identifiers in the input features and achieve an accurate and efficient representation of omics data when dealing with complex multi-omics data.

Subtype-Former has limitations. Firstly, due to the peculiarity of deep learning, Subtype-Former needs sufficient training data to support it to achieve optimal cancer subtyping performance. If the training samples are rare, Subtype-Former may reach a local optimum, resulting in significant performance degradation. Our future study direction is how to develop unsupervised few-shot learning methods to improve subtyping accuracy on rare cancers that lack sufficient samples. In addition, missing data can also impact the performance of the Subtype-Former. We currently deal with missing data by simply filling in the mean values and not considering the correlation between multi-omics data. We plan to develop a GAN method to efficiently fix the missing data and improve cancer subtyping accuracy in our future work.

## Funding

*This work is supported by Natural Science Foundation of China under Grant No. 61902126, Shanghai Science and Technology Program "Distributed and generative few-shot algorithm and theory research" under Grant No. 20511100600, Shanghai Science and Technology Program "Federated based cross-domain and cross-task incremental learning" under Grant No. 21511100800*

*Conflict of Interest:* none declared.

## References

Abeshouse,A. *et al.* (2015) The Molecular Taxonomy of Primary Prostate Cancer. *Cell*, **163**, 1011–1025.
Adamson,C. *et al.* (2009) Glioblastoma multiforme: a review of where we have been and where we are going. *Expert Opin. Investig. Drugs*, **18**, 1061–1083.
Agrawal,N. *et al.* (2014) Integrated Genomic Characterization of Papillary Thyroid Carcinoma. *Cell*, **159**, 676–690.
Akbani,R. *et al.* (2015) Genomic Classification of Cutaneous Melanoma. *Cell*, **161**, 1681–1696.
Berger,A.C. *et al.* (2018) A Comprehensive Pan-Cancer Molecular Study of Gynecologic and Breast Cancers. *Cancer Cell*, **33**, 690-705.e9.
Chauvel,C. *et al.* (2020) Evaluation of integrative clustering methods for the analysis of multi-omics data. *Brief. Bioinform.*, **21**, 541–552.
Chung,C.H. *et al.* (2004) Molecular classification of head and neck squamous cell carcinomas using patterns of gene expression. *Cancer Cell*, **5**, 489–500.
Dasgupta,P. *et al.* (2020) LncRNA CDKN2B-AS1/miR-141/cyclin D network regulates tumor progression and metastasis of renal cell carcinoma. *Cell Death Dis.*, **11**, 660.
Duan,R. *et al.* (2021) Evaluation and comparison of multi-omics data integration methods for cancer subtyping. *PLOS Comput. Biol.*, **17**, e1009224.
Hoadley,K.A. *et al.* (2018) Cell-of-Origin Patterns Dominate the Molecular Classification of 10,000 Tumors from 33 Types of Cancer. *Cell*, **173**, 291-304.e6.
Li,Y. *et al.* (2016) A review on machine learning principles for multi-view biological data integration. *Brief. Bioinform.*, bbw113.
Liu,M. *et al.* (2021) A method for subtype analysis with somatic mutations. *Bioinformatics*, **37**, 50–56.
Liu,Y. *et al.* (2018) Comparative Molecular Analysis of Gastrointestinal Adenocarcinomas. *Cancer Cell*, **33**, 721-735.e8.
Liu,Z. *et al.* (2020) Identification of methylation-driven genes related to the prognosis of papillary renal cell carcinoma: a study based on The Cancer Genome Atlas. *Cancer Cell Int.*, **20**, 235.
Lu,X. *et al.* (2021) *MOVICS* : an R package for multi-omics integration and visualization in cancer subtyping. *Bioinformatics*, **36**, 5539–5541.
Meng,C. *et al.* (2016) moCluster: Identifying Joint Patterns Across Multiple Omics Data Sets. *J. Proteome Res.*, **15**, 755–765.
Meng,C. *et al.* (2019) MOGSA: Integrative Single Sample Gene-set Analysis of Multiple Omics Data. *Mol. Cell. Proteomics*, **18**, S153–S168.
Mo,Q. *et al.* (2018) A fully Bayesian latent variable model for integrative clustering analysis of multi-type omics data. *Biostatistics*, **19**, 71–86.
Nguyen,H. *et al.* (2019) PINSPlus: a tool for tumor subtype discovery in integrated genomic data. *Bioinformatics*, **35**, 2843–2846.
Nguyen,T. *et al.* (2017) A novel approach for data integration and disease subtyping. *Genome Res.*, **27**, 2025–2039.
Pleasance,E. *et al.* (2020) Pan-cancer analysis of advanced patient tumors reveals interactions between therapy and genomic landscapes. *Nat. Cancer*, **1**, 452–468.
Ramazzotti,D. *et al.* (2018) Multi-omic tumor data reveal diversity of molecular mechanisms that correlate with survival. *Nat. Commun.*, **9**, 4453.
Rappoport,N. and Shamir,R. (2018) Multi-omic and multi-view clustering algorithms: review and cancer benchmark. *Nucleic Acids Res.*, **46**, 10546–10562.
Rappoport,N. and Shamir,R. (2019) NEMO: cancer subtyping by integration of partial multi-omic data. *Bioinformatics*, **35**, 3348–3356.
Reed,S. *et al.* (2022) A Generalist Agent.
Song,W. *et al.* (2022) Subtype-WESLR: identifying cancer subtype with weighted ensemble sparse latent representation of multi-view data. *Brief. Bioinform.*, **23**, bbab398.
The Cancer Genome Atlas Research Network (2012) Comprehensive genomic characterization of squamous cell lung cancers. *Nature*, **489**, 519–525.
The Cancer Genome Atlas Research Network (2013) Comprehensive molecular characterization of clear cell renal cell carcinoma. *Nature*, **499**, 43–49.
The Cancer Genome Atlas Research Network (2014) Comprehensive molecular profiling of lung adenocarcinoma. *Nature*, **511**, 543–550.
The Cancer Genome Atlas Research Network *et al.* (2013) The Cancer Genome Atlas Pan-Cancer analysis project. *Nat. Genet.*, **45**, 1113–1120.
The Cancer Genome Atlas Research Network and Levine,D.A. (2013) Integrated genomic characterization of endometrial carcinoma. *Nature*, **497**, 67–73.
The International Cancer Genome Consortium (2010) International network of cancer genome projects. *Nature*, **464**, 993–998.
Thosani,S. *et al.* (2013) The Characterization of Pheochromocytoma and Its Impact on Overall Survival in Multiple Endocrine Neoplasia Type 2. *J. Clin. Endocrinol. Metab.*, **98**, E1813–E1819.
Tunyasuvunakool,K. *et al.* (2021) Highly accurate protein structure prediction for the human proteome. *Nature*, **596**, 590–596.
Ushijima,T. *et al.* (2021) Mapping genomic and epigenomic evolution in cancer ecosystems. *Science*, **373**, 1474–1479.
Vitale,I. *et al.* (2021) Intratumoral heterogeneity in cancer progression and response to immunotherapy. *Nat. Med.*, **27**, 212–224.
Wang,B. *et al.* (2014) Similarity network fusion for aggregating data types on a genomic scale. *Nat. Methods*, **11**, 333–337.

Witten,D.M. and Tibshirani,R.J. (2009) Extensions of Sparse Canonical Correlation Analysis with Applications to Genomic Data. *Stat. Appl. Genet. Mol. Biol.*, **8**, 1–27.

Wu,D. *et al.* (2015) Fast dimension reduction and integrative clustering of multi-omics data using low-rank approximation: application to cancer molecular classification. *BMC Genomics*, **16**, 1022.

Xiao,W. *et al.* (2019) HAO2 inhibits malignancy of clear cell renal cell carcinoma by promoting lipid catabolic process. *J. Cell. Physiol.*, **234**, 23005–23016.

Yang,H. *et al.* (2021) Subtype-GAN: a deep learning approach for integrative cancer subtyping of multi-omics data. *Bioinformatics*, **37**, 2231–2237.

Zhao,L. *et al.* (2019) Molecular subtyping of cancer: current status and moving toward clinical applications. *Brief. Bioinform.*, **20**, 572–584.